\documentclass[11pt,letterpaper]{article}
\usepackage[nohyperref]{emnlp2017}
\usepackage{emnlp2017}
\usepackage{times}
\usepackage{latexsym}
\usepackage{booktabs}
\usepackage{graphicx,url,color}
\usepackage{amsmath,amssymb,amsopn,amsthm,bm,bbm}
\usepackage{algorithm,algpseudocode,algorithmicx}
\usepackage{verbatim}
\usepackage[margin=1.2in]{geometry}
\usepackage[disable]{todonotes}
\usepackage{multirow,rotating}
\usepackage[normalem]{ulem}
\usepackage{natbib}
\usepackage{placeins}
\usepackage[procnames]{listings}

\emnlpfinalcopy

\def\emnlppaperid{637}

\newcommand{\Y}{\mathcal{Y}}
\newcommand{\X}{\mathcal{X}}
\newcommand{\E}{\mathbb{E}}

\newcommand{\D}{\mathcal{D}}

\newcommand{\cDC}{$\hat{c}$\,DC}
\newcommand{\cDR}{$\hat{c}$\,DR}

\title{Counterfactual Learning from Bandit Feedback under Deterministic Logging: A Case Study in Statistical Machine Translation}
\author{Carolin Lawrence \\
  Heidelberg University,\\Germany \\\And
  Artem Sokolov \\
  Amazon Development Center \\ \& Heidelberg University, Germany\\
  {\small \tt \{lawrence,sokolov,riezler\}@cl.uni-heidelberg.de} \\\And
  Stefan Riezler\\
  Heidelberg University,\\Germany}

\date{}

\begin{document}

\maketitle

\begin{abstract}
 The goal of counterfactual learning for statistical machine translation (SMT) is to optimize a target SMT system from logged data that consist of user feedback to translations that were predicted by another, historic SMT system. A challenge arises by the fact that risk-averse commercial SMT systems deterministically log the most probable translation. The lack of sufficient exploration of the SMT output space seemingly contradicts the theoretical requirements for counterfactual learning. We show that counterfactual learning from deterministic bandit logs is possible nevertheless by smoothing out deterministic components in learning. This can be achieved by additive and multiplicative control variates that avoid degenerate behavior in empirical risk minimization. Our simulation experiments show improvements of up to 2 BLEU points by counterfactual learning from deterministic bandit feedback.
\end{abstract}

\section{Introduction}

\begin{figure}[h!] 
	\centering
	\includegraphics[width=0.25\textwidth,keepaspectratio]{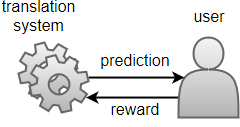}
	\caption{Online learning from partial feedback.}
	\label{fig:diagram_on_policy}
\end{figure}

Commercial SMT systems allow to record large amounts of interaction log data at no cost. Such logs typically contain a record of the source, the translation predicted by the system, and the user feedback. The latter can be gathered directly if explicit user quality ratings of translations are supported, or inferred indirectly from the interaction of the user with the translated content. Indirect feedback in form user clicks on displayed ads has been shown to be a valuable feedback signal in response prediction for display advertising \cite{BottouETAL:13}. Similar to the computational advertising scenario, one could imagine a scenario where SMT systems are optimized from partial information in form of user feedback to predicted translations, instead of from manually created reference translations. This learning scenario has been investigated in the areas of \emph{bandit learning} \cite{BubeckCesaBianchi:12} or \emph{reinforcement learning} (RL) \cite{SuttonBarto:98}. Figure \ref{fig:diagram_on_policy} illustrates the learning protocol using the terminology of \emph{bandit structured prediction} \cite{SokolovETALnips:16,KreutzerETAL:17}, where at each round, a \emph{system} (corresponding to a \emph{policy} in RL terms) makes a prediction (also called \emph{action} in RL, or pulling an \emph{arm} of a bandit), and receives a \emph{reward}, which is used to update the system.

\begin{figure}[h!]
	\centering
	\includegraphics[width=0.25\textwidth,keepaspectratio]{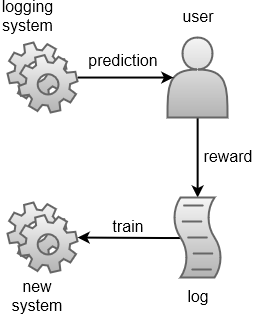}
        \caption{Offline learning from partial feedback.}
	\label{fig:diagram_off_policy}
\end{figure}

\emph{Counterfactual} learning attempts to reuse existing interaction data where the predictions have been made by a historic system different from the target system. This enables \emph{offline} or \emph{batch} learning from logged data, and is important if online experiments that deploy the target system are risky and/or expensive.
Counterfactual learning tasks include \emph{policy evaluation}, i.e. estimating how a target policy would have performed if it had been in control of choosing the predictions for which the rewards were logged, and \emph{policy optimization} (also called \emph{policy learning}), i.e. optimizing parameters of a target policy given the logged data from the historic system. Both tasks are called \emph{counterfactual}, or \emph{off-policy} in RL terms, since the target policy was actually not in control during logging. Figure \ref{fig:diagram_off_policy} shows the learning protocol for off-policy learning from partial feedback.

The crucial trick to obtain unbiased estimators to evaluate and to optimize the off-policy system is to correct the sampling bias of the logging policy. This can be done by importance sampling where the estimate is corrected by the inverse propensity score  \cite{RosenbaumRubin:83} of the historical algorithm, mitigating the problem that predictions there were favored by the historical system are over-represented in the logs. As shown by \citet{LangfordETAL:08} or \citet{StrehlETAL:10}, a sufficient exploration of the output space by the logging system is a prerequisite for counterfactual learning. If the logging policy acts stochastically in predicting outputs, this condition is satisfied, and inverse propensity scoring can be applied to correct the sampling bias. However, commercial SMT systems usually try to avoid any risk and only log the most probable translation. This effectively results in deterministic logging policies, making theory and practice of off-policy methods inapplicable to counterfactual learning in SMT. 

This paper presents a case study in counterfactual learning for SMT that shows that policy optimization from deterministic bandit logs is possible despite these seemingly contradictory theoretical requirements. We formalize our learning problem as an empirical risk minimization over logged data. While a simple empirical risk minimizer can show degenerate behavior where the objective is minimized by avoiding or over-representing training samples, thus suffering from decreased generalization ability, we show that the use of control variates can remedy this problem. Techniques such as doubly-robust policy evaluation and learning \citep{DudikETAL:11} or weighted importance sampling \citep{JiangLi:16,ThomasBrunskill:16} can be interpreted as additive \citep{Ross:13} or multiplicative control variates \citep{Kong:92} that serve for variance reduction in estimation. We observe that a further effect of these techniques is that of smoothing out deterministic components by taking the whole output space into account. Furthermore, we conjecture that while outputs are logged deterministically, the stochastic selection of inputs serves as sufficient exploration in parameter optimization over a joint feature representation over inputs and outputs. We present experiments using simulated bandit feedback for two different SMT tasks, showing improvements of up to 2 BLEU in SMT domain adaptation from deterministically logged bandit feedback. This result, together with a comparison to the standard case of policy learning from stochastically logged simulated bandit feedback, confirms the effectiveness our proposed techniques.

\section{Related Work}

Counterfactual learning has been known under the name of off-policy learning in various fields that deal with partial feedback, namely contextual bandits (\citet{LangfordETAL:08,StrehlETAL:10,DudikETAL:11,LiETAL:15}, \emph{inter alia}), reinforcement learning (\citet{SuttonBarto:98,PrecupETAL:00,JiangLi:16,ThomasBrunskill:16}, \emph{inter alia}), and structured prediction (\citet{SwaminathanJoachimsJMLR:15,SwaminathanJoachimsNIPS:15}, \emph{inter alia}). The idea behind these approaches is to first perform policy evaluation and then policy optimization, under the assumption that better evaluation leads to better optimization. Our work puts a focus on policy optimization in an empirical risk minimization framework for deterministically logged data. Since our experiment is a simulation study, we can compare the deterministic case to the standard scenario of policy optimization and evaluation under stochastic logging.

Variance reduction by additive control variates has implicitly been used in doubly robust techniques \citep{DudikETAL:11,JiangLi:16}. However, the connection to Monte Carlo techniques has not been made explicit until \citet{ThomasBrunskill:16}, nor has the control variate technique of optimizing the variance reduction by adjusting a linear interpolation scalar \citep{Ross:13} been applied in off-policy learning. Similarly, the technique of weighted importance sampling has been used as variance reduction technique in off-policy learning \citep{PrecupETAL:00,JiangLi:16,ThomasBrunskill:16}. The connection to multiplicative control variates \citep{Kong:92} has been made explicit in \citet{SwaminathanJoachimsNIPS:15}. To our knowledge, our analysis of both control variate techniques from the perspective of avoiding degenerate behavior in learning from deterministically logged data is novel.

\section{Counterfactual Learning from Deterministic Bandit Logs}

\begin{table*}[t]
	\begin{center}
		\begin{tabular}{l}
			\toprule
                        $\nabla \hat{R}_{\text{DPM}} =  \frac{1}{n} \sum_{t=1}^{n} \delta_t \pi_w(y_t | x_t) \nabla \log \pi_w(y_t | x_t).$ \\ \midrule
                        $\nabla\hat{R}_{\text{DPM+R}} = \frac{1}{n} \sum_{t=1}^{n} [ \delta_t \bar{\pi}_w(y_t|x_t) (  \nabla \log \pi_w(y_t | x_t)  - \sum_{u=1}^{n} \bar{\pi}_w(y_u|x_u)  \nabla \log \pi_w(y_u | x_u) )].$ \\ \midrule
                        $\nabla\hat{R}_{\text{\cDC}} = \frac{1}{n} \sum_{t=1}^{n} [ (\delta_t-\hat{c}\hat{\delta}_t) \bar{\pi}_w(y_t|x_t) (  \nabla \log \pi_w(y_t | x_t) - \sum_{u=1}^{n} \bar{\pi}_w(y_u|x_u)  \nabla \log \pi_w(y_u | x_u)) $ \\
                          $ \qquad \qquad \qquad + \hat{c} \sum_{y \in \mathcal{Y}(x_t)} \hat{\delta}(x_t,y)  \pi_w(y | x_t) \nabla \log \pi_w(y | x_t)].$ \\
			\bottomrule
		\end{tabular}
		\caption{Gradients of counterfactual objectives.}
                \label{tab:gradients}
	\end{center}
\end{table*}
\paragraph{Problem Definition.}

The problem of counterfactual learning (in the following used in the sense of counterfactual optimization) for bandit structured prediction can be described as follows: Let $\X$ be a structured input space, let $\Y(x)$ be the set of possible output structures for input $x$, and let $\Delta: \Y  \rightarrow [0,1]$ be a reward function (and $\delta = - \Delta$ be the corresponding task loss function)\footnote{We will use both terms, reward and loss, in order to be consistent with the respective literature.} quantifying the quality of structured outputs. We are given a data log of triples $\D=\{(x_t,y_t,\delta_t)\}_{t=1}^n$ where outputs $y_t$ for inputs $x_t$ were generated by a logging system, and loss values $\delta_t$ were observed only at the generated data points. In case of stochastic logging with probability $\pi_0$, the inverse propensity scoring approach \citep{RosenbaumRubin:83} uses importance sampling to achieve an unbiased estimate of the expected loss under the parametric target policy $\pi_w$:
\begin{align}
  \label{eq:ips}
\hat{R}_{\text{IPS}}(\pi_w) & = \frac{1}{n} \sum_{t=1}^n \delta_t \frac{\pi_w(y_t|x_t)}{\pi_0(y_t|x_t)} \\ \notag
& \approx \E_{p(x)} \E_{\pi_0(y|x)} [\delta(y) \frac{\pi_w(y|x)}{\pi_0(y|x)}] \\ \notag
& =  \E_{p(x)} \E_{\pi_w(y|x)} [\delta(y)]. \\ \notag
\end{align}
\vspace{-5ex}

In case of deterministic logging, we are confined to empirical risk minimization:
\begin{align}
\label{eq:emp-risk} 
\hat{R}_{\text{DPM}}(\pi_w) = \frac{1}{n} \sum_{t=1}^n \delta_t \pi_w(y_t|x_t).
\end{align}
Equation \eqref{eq:emp-risk} assumes deterministically logged outputs with propensity $\pi_0 =1, t=1, \ldots, n$ of the historical system. We call this objective the \emph{deterministic propensity matching (DPM)} objective since it matches deterministic outputs of the logging system to outputs in the $n$-best list of the target system. For optimization under deterministic logging, a sampling bias is unavoidable since objective \eqref{eq:emp-risk} does not correct it by importance sampling. Furthermore, the DPM estimator may show a degenerate behavior in learning. This problem can be remedied by the use of control variates, as we will discuss in Section \ref{sec:analysis}.

\paragraph{Learning Principle: Doubly Controlled Empirical Risk Minimization.}

Our first modification of Equation \eqref{eq:emp-risk} has been originally motivated by the use of weighted importance sampling in inverse propensity scoring because of its observed stability and variance reduction effects \citep{PrecupETAL:00,JiangLi:16,ThomasBrunskill:16}. We call this objective the \emph{reweighted deterministic propensity matching (DPM+R)} objective:
\begin{align}
\label{eq:r-risk} 
\hat{R}_{\text{DPM+R}}(\pi_w) &= \frac{1}{n}\sum_{t=1}^n \delta_t \bar{\pi}_w(y_t|x_t) \\ \notag
&= \frac{1}{n}\sum_{t=1}^n \delta_t \frac{\pi_w(y_t|x_t)}{\frac{1}{n}\sum_{t=1}^n \pi_w(y_t|x_t)}.\\ \notag
\end{align}
\vspace{-5ex}

From the perspective of Monte Carlo simulation, the advantage of this modification can be explained by viewing reweighting as a multiplicative control variate \citep{SwaminathanJoachimsNIPS:15}. Let $Z = \delta_t \pi_w(y_t|x_t)$ and $W = \pi_w(y_t|x_t)$ be two random variables, then the variance of $r =\frac{\frac{1}{n}\sum_{t=1}^n Z}{\frac{1}{n}\sum_{t=1}^n W}$ can be approximately written as follows \citep{Kong:92}: $\text{Var}(r) \approx \frac{1}{n} (r^2 \text{Var}(W) + \text{Var}(Z) - 2r\,\text{Cov}(W,Z)).$
This shows that a positive correlation between the variable $W$, representing the target model probability, and the variable $Z$, representing the target model scaled by the task loss function, will reduce the variance of the estimator. Since there are exponentially many outputs to choose from for each input during logging, variance reduction is useful in counterfactual learning even in the deterministic case. Under a stochastic logging policy, a similar modification can be done to objective \eqref{eq:ips} by reweighting the ratio $\rho_t = \frac{\pi_w(y_t|x_t)}{\pi_0(y_t|x_t)}$ as $\bar{\rho_t} = \frac{\rho_t}{\sum_t \rho_t}$. We will use this reweighted IPS objective, called IPS+R, in our comparison experiments that use stochastically logged data.

A further modification of Equation \eqref{eq:r-risk} is motivated by the incorporation of a direct reward estimation method in the inverse propensity scorer as proposed in the doubly-robust estimator \citep{DudikETAL:11,JiangLi:16,ThomasBrunskill:16}. Let $\hat{\delta}(x_t, y_t)$ be a regression-based reward model trained on the logged data, and let $\hat{c}$ be a scalar that allows to optimize the estimator for minimal variance \cite{Ross:13}. We define a \emph{doubly controlled} empirical risk minimization objective $\hat{R}_{\text{\cDC}}$ as follows (for $\hat{c} = 1$ we arrive at a similar objective called $\hat{R}_{\text{DC}}$):
\begin{align}
\label{eq:dr-risk} 
\hat{R}_{\text{\cDC}}(\pi_w) =& \frac{1}{n} \sum_{t=1}^{n} \Big[  (\delta_t-\hat{c}\hat{\delta_t}) \; \bar{\pi}_w(y_t | x_t) \\ \notag
&+  \hat{c} \sum_{y \in \mathcal{Y}(x_t)} \hat{\delta}(x_t,y) \; \pi_w(y | x_t) \Big]. \\ \notag
\end{align}
\vspace{-5ex}
 
From the perspective of Monte Carlo simulation, the doubly robust estimator can be seen as variance reduction via additive control variates \citep{Ross:13}. Let $X =\delta_t$ and $Y = \hat{\delta}_t$ be two random variables. Then $\bar{Y} = \sum_{y \in \mathcal{Y}(x_t)} \hat{\delta}(x_t,y) \; {\pi}_w(y | x_t)$ is the expectation\footnote{Note that we introduce a slight bias by using $\pi_w$ versus $\bar{\pi}_w$ in sampling probability and control variate.} of $Y$, and Equation \eqref{eq:dr-risk} can be rewritten as $\mathbb{E}_{\bar{\pi}_w(x)}(X-\hat{c}\,Y) + \hat{c}\,\bar{Y}.$
The variance of the term $X-\hat{c}\,Y$ is $\text{Var}(X-\hat{c}\,Y) =  \text{Var}(X) + \hat{c}^2\text{Var}(Y) - 2\hat{c}\,\text{Cov}(X,Y).$ (\citet{Ross:13}, Chap. 9.2).
Again this shows that variance of the estimator can be reduced if the variable $X$, representing the reward function, and the variable $Y$, representing the regression-based reward model, are positively correlated. The optimal scalar parameter $\hat{c}$ can be derived easily by taking the derivative of variance term, leading to
\begin{align}
\label{eq:optimal-control-variate} 
\hat{c} = \frac{\text{Cov}(X,Y)}{\text{Var}(Y)}.
\end{align}
In case of stochastic logging the reweighted target probability $\bar{\pi}_w(y_t|x_t)$ is replaced by a reweighted ratio $\bar{\rho_t}$. We will use such reweighted models of the original doubly robust model, with and without optimal $\hat{c}$, called DR and \cDR{}, in our experiments that use stochastic logging.

\paragraph{Learning Algorithms.}

Applying a stochastic gradient descent update rule $w_{t+1} = w_t - \eta \nabla\hat{R}(\pi_w)_t$
to the objective functions defined above leads to a variety of algorithms. The gradients of the objectives can be derived by using the score function gradient estimator \citep{Fu:06} and are shown in Table \ref{tab:gradients}. Stochastic gradient descent algorithms apply to any differentiable policy $\pi_w$, thus our methods can be applied to a variety of systems, including linear and non-linear models. Since previous work on off-policy methods in RL and contextual bandits has been done in the area of linear classification, we start with an adaptation of off-policy methods to linear SMT models in our work. We assume a Gibbs model
\begin{align}
  \label{eq:gibbs}
  \pi_w(y_t | x_t) = \frac{e^{\alpha (w^{\top} \phi(x_t,y_t))}}{\sum_{y\in \mathcal{Y}(x_t)} e^{\alpha (w^{\top} \phi(x_t, y))}},
\end{align}
based on a feature representation $\phi:\X \times \Y \rightarrow \mathbb{R}^d$, a weight vector $w \in \mathbb{R}^d$, and a smoothing parameter $\alpha \in \mathbb{R}^{+}$, yielding the following simple derivative $\nabla \log \pi_w(y_t | x_t) = \alpha \big(\phi(x_t,y_t) - \sum_{y\in \mathcal{Y}(x_t)}  \phi(x_t, y)\pi_w(y_t | x_t)\big).$

\section{Experiments}

\paragraph{Setup.} In our experiments, we aim to simulate the following scenario: We assume that it is possible to divert a small fraction of the user interaction traffic for the purpose of policy evaluation and to perform stochastic logging on this small data set. The main traffic is assumed to be logged deterministically, following a conservative regime where one-best translations are used for an SMT system that does not change frequently over time. Since our experiments are simulation studies, we will additionally perform stochastic logging, and compare policy learning for the (realistic) case of deterministic logging with the (theoretically motivated) case of stochastic logging.

In our deterministic-based policy learning experiments, we evaluate the empirical risk minimization algorithms derived from objectives \eqref{eq:r-risk} (DPM+R) and \eqref{eq:dr-risk}. For the doubly controlled objective we employ two variants: First, $\hat{c}$ is set to 1 as in \cite{DudikETAL:11} (DC). Second, we calculate $\hat{c}$ as described in Equation \eqref{eq:optimal-control-variate} (\cDC{}). The algorithms used in policy evaluation and for stochastic-based policy learning are variants of these objectives that replace $\bar{\pi}$ by $\bar{\rho}$ to yield estimators IPS+R, DR, and {\cDR{}} of the expected loss.

\begin{table}
	\begin{center}
		\begin{tabular}{c|cc}
			& TED DE-EN & News FR-EN \\
			\toprule
			train         & 122k & 30k \\
			validation& 3k & 1k \\
			test& 3k & 2k \\
			\bottomrule
		\end{tabular}
		\caption{Number of sentences for in-domain data splits of SMT train, validation, and test data.}
		\label{exp:data_splits}
	\end{center}
\end{table}

All objectives will be employed in a domain adaptation scenario for machine translation. A system trained on out-of-domain data will be used to collect feedback on in-domain data. This data will serve as the logged data $\D$ in the learning experiments.

We conduct two SMT tasks with hypergraph re-decoding: The first is German-to-English and is trained using a concatenation of the Europarl corpus \citep{europarl}, the Common Crawl corpus\footnote{\url{http://www.statmt.org/wmt13/training-parallel-commoncrawl.tgz}} and the News Commentary corpus \citep{KoehnSchroeder:07}. The goal is to adapt the trained system to the domain of transcribed TED talks using the TED parallel corpus \citep{ted}. A second task uses the French-to-English Europarl data with the goal of domain adaptation to news articles with the News Commentary corpus \citep{KoehnSchroeder:07}. We split off two parts from the TED corpus to be used as validation and test data for the learning experiments. As validation data for the News Commentary corpus we use the splits provided at the WMT shared task, namely \texttt{nc-devtest2007} as validation data and \texttt{nc-test2007} as test data. An overview of the data statistics can be seen in Table \ref{exp:data_splits}.

As baseline, an out-of-domain system is built using the SCFG framework \textsc{cdec} \citep{DyerETAL:2010} with dense features (10 standard features and 2 for the language model). After tokenizing and lowercasing the training data, the data were word aligned using \textsc{cdec}'s \texttt{fast\_align}. A 4-gram language model is build on the target languages for the out-of-domain data using \textsc{KenLM} \citep{HeafieldETAL:13}. For News, we additionally assume access to in-domain target language text and train another in-domain language model on that data, increasing the number of features to 14 for News.

The framework uses a standard linear Gibbs model whose distribution can be peaked using a parameter $\alpha$ (see Equation \eqref{eq:gibbs}): Higher value of $\alpha$ will shift the probability of the one-best translation closer to 1 and all others closer to 0. Using $\alpha > 1$ during training will promote to learn models that are optimal when outputting the one-best translation. In our experiments, we found $\alpha=5$ to work well on validation data.

Additionally, we tune a system using \textsc{cdec}'s MERT implementation \citep{Och:03} on the in-domain data with their references. This full-information in-domain system conveys the best possible improvement using the given training data. It can thus be seen as the oracle system for the systems which are learnt using the same input-side training data, but have only bandit feedback available to them as a learning signal. All systems are evaluated using the corpus-level BLEU metric \citep{PapineniETAL:02}.

The logged data $\D$ is created by translating the in-domain training data of the corpora using the original out-of-domain systems, and logging the one-best translation. For the stochastic experiments, the translations are sampled from the model distribution. The feedback to the logged translation is simulated using the reference and sentence-level BLEU \citep{NakovETAL:12}.

\paragraph{Direct Reward Estimation.}

When creating the logged data $\D$, we also record the feature vectors of the translations to train the direct reward estimate that is needed for ($\hat{c}$)DC. Using the feature vector as input and the per-sentence BLEU as the output value, we train a regression-based random forest with 10 trees using scikit-learn \citep{scikit-learn}. To measure performance, we perform 5-fold cross-validation and measure the macro average between estimated rewards and the true rewards from the log: $|\frac{1}{n} \sum \delta(x_t,y_t) - \frac{1}{n} \sum \hat{\delta}(x_t,y_t)|$. 
We also report the micro average which quantifies how far off one can expect the model to be for a random sample: $\frac{1}{n} \sum |\delta(x_t,y_t) - \hat{\delta}(x_t,y_t) |$. 
The final model used in the experiments is trained on the full training data. Cross-validation results for the regression-based direct reward model can be found in Table~\ref{dm_results}.

\begin{table}
	\centering
	\begin{tabular}{l|ll}
		&TED&News\\
		\toprule
		macro avg.&0.67&0.23\\
		micro avg.&15.03&10.87\\
		\bottomrule
	\end{tabular}
	\caption{Evaluation of regression-based reward estimation by average BLEU differences between estimated and true rewards.}
	\label{dm_results}
\end{table}

\begin{table}
	\centering
	\normalsize
	\begin{tabular}{ll|ccc}
		&&IPS+R&DR&\cDR{}\\
		\toprule
		\multirow{2}{*}{\begin{sideways}TED\end{sideways}}&avg. estimate&+4.00&+7.98&+6.07\\
		&std. dev.&0.64&3.83&2.06\\
		\midrule
		\multirow{2}{*}{\begin{sideways}News\end{sideways}}&avg. estimate&-7.78&+6.63&+0.95\\
		&std. dev.&0.97&4.13&2.33\\
		\bottomrule
	\end{tabular}\caption{Policy evaluation by macro averaged difference between estimated and ground truth BLEU on 10k stochastically logged data, averaged over 5 runs.} 
	\label{tab:eval}
\end{table}

\paragraph{Policy Evaluation.}

Policy evaluation aims to use the logged data $\D$ to estimate the performance of the target system $\pi_w$. The small logged data $\D_{eval}$ that is diverted for policy evaluation is created by translating only 10k sentences of the in-domain training data with the out-of-domain system and sample translations according to the model probability. Again we record the sentence-level BLEU as the feedback. The reference translations that also exist for those 10k sentences are used to measure the ground truth BLEU value for translations using the full-information in-domain system. The goal of evaluation is to achieve a value of IPS+R, DR, and \cDR{} on $\D_{eval}$ that are as close as possible to the ground truth BLEU value.

To be able to measure variance, we create five folds of $\D_{eval}$, differing in random seeds. We report the average difference between the ground truth BLEU score and the value of the log-based policy evaluation, as well as the standard deviation in Table \ref{tab:eval}. We see that IPS+R underestimates the BLEU value by 7.78 on News. DR overestimates instead. \cDR{} achieves the closest estimate, overestimating the true value by less than 1 BLEU. On TED, all policy evaluation results are overestimates. For the DR variants the overestimation result can be explained by the random forests' tendency to overestimate. Optimal \cDR{} can correct for this, but not always in a sufficient way.

\begin{table*}[!htbp]
	\centering
	\begin{tabular}{lll|ccccc}
		&&&BLEU&\multicolumn{3}{c}{BLEU difference}&BLEU\\
		&&&out-of-domain&DPM+R&DC&\cDC{}&in-domain\\
		\toprule
		\multirow{4}{*}{\begin{sideways}deterministic\end{sideways}}&\multirow{2}{*}{\begin{sideways}TED\end{sideways}}&validation&22.39&+0.59&+1.50&+1.89&25.43\\
		&&test&22.76&+0.67&+1.41&+2.02&25.58\\
		\cmidrule{2-8}
		&\multirow{2}{*}{\begin{sideways}News\end{sideways}}&validation&24.64&+0.62&+0.99&+1.02&27.62\\
		&&test&25.27&+0.94&+1.05&+1.13&28.08\\
		\midrule
		\midrule
		&&&out-of-domain&IPS+R&DR&\cDR{}&in-domain\\
		\midrule
		\multirow{4}{*}{\begin{sideways}\hskip -0.3cm stochastic\end{sideways}}&\multirow{2}{*}{\begin{sideways}TED\end{sideways}}&validation&22.39&+0.57&+1.92&+1.95&25.43\\
		&&test&22.76&+0.58&+2.04&+2.09&25.58\\
		\cmidrule{2-8}
		&\multirow{2}{*}{\begin{sideways}News\end{sideways}}&validation&24.64&+0.71&+1.00&+0.71&27.62\\
		&&test&25.27&+0.81&+1.18&+0.95&28.08\\
		\bottomrule
	\end{tabular}\caption{BLEU increases for learning, over the out-of-domain baseline on validation and test set. Out-of-domain is the baseline and starting system and in-domain is the oracle system tuned on in-domain data with references. For the deterministic case, all results are statistically significant at $p \le 0.001$ with regards to the baseline. For the stochastic case, all results are statistically significant at $p \le 0.002$ with regards to the baseline, except for IPS+R on the News corpus.} 
	\label{exp:learn_results}
\end{table*}

\paragraph{Policy Learning.}

In our learning experiments, learning starts with the weights $w_0$ from the out-of-domain model. As this was the system that produced the logged data $\D$, the first iteration will have the same translations in the one-best position. After some iterations, however, the translation that was logged may not be in the first position any more. In this case, the $n$-best list is searched for the correct translation. Due to speed reasons, the scores of the translation system are normalized to probabilities using the first 1,000 unique entries in the $n$-best list, rather than using the full hypergraph. Our experiments showed that this did not impact the quality of learning.

In order for the multiplicative control variate to be effective, the learning procedure has to utilize mini-batches. If the mini-batch size is chosen too small, the estimates of the control variates may not be reliable. We test mini-batch sizes of 30k and 10k examples, whereas 30k on News means that we perform batch training since the mini-batch spans the entire training set. Mini-batch size $\beta$ and early stopping point where selected by choosing the setup and iteration that achieved the highest BLEU score on the one-best translations for the validation data. The learning rate $\eta$ was selected in the same way, whereas the possible values were {$1\mathrm{e}{-4}, 1\mathrm{e}{-5}, 1\mathrm{e}{-6}$} or, alternatively, Adadelta \citep{Zeiler:12}, which sets the learning rate on a per-feature basis.
The results on both validation and test set are reported in Table \ref{exp:learn_results}. Statistical significance of the out-of-domain system compared to all other systems is measured using Approximate Randomization testing \citep{Noreen:1989}.

For the deterministic case, we see that in general DPM+R shows the lowest increase but can still significantly outperform the baseline. An explanation of why DPM+R cannot improve any further, will be addressed separately below. DC yields improvements of up to 1.5 BLEU points, while \cDC{} obtains improvements of up to 2 BLEU points over the out-of-domain baseline.
In more detail on the TED data, DC can close the gap of nearly 3 BLEU by half between the out-of-domain and the full-information in-domain system. \cDC{} can improve by further 0.6 BLEU which is a significant improvement at $p=0.0017$. Also note that, while \cDC{} takes more iterations to reach its best result on the validation data, \cDC{} already outperforms DC at the stopping iteration of DC. At this point \cDC{} is better by 0.18 BLEU on the validation set and continues to increase until its own stopping iteration. The final results of \cDC{} falls only 0.8 BLEU behind the oracle system that had references available during its learning process. Considering the substantial difference in information that both systems had available, this is remarkable.
The improvements on the News corpus show similar tendencies. Again there is a gap of nearly 3 BLEU to close and with an improvement of 1.05 BLEU points, DC can achieve a notable result. \cDC{} was able to further improve on this but not as successfully as was the case for the TED corpus. Analyzing the actual $\hat{c}$ values that were calculated in both experiments allows us to gain an insight as to why this was the case: For TED, $\hat{c}$ is on average 1.35. In the case of News, however, $\hat{c}$ has a maximum value of 1.14 and thus stays quite close to 1, which would equate to using DC. It is thus not surprising that there is no significant difference between DC and \cDC{}.

\paragraph{Comparison to the Stochastic Case.} Even if not realistic for commercial applications of SMT, our simulation study allows us to stochastically log large amounts of data in order to compare learning from deterministic logs to the standard case. As shown in Table \ref{exp:learn_results}, the relations between algorithms and even the absolute improvements are similar for stochastic and deterministic logging. Significance tests between each deterministic/stochastic experiment pair show a significant difference only in case of DC/DR on TED data. However, the DR result still does not significantly outperform the best deterministic objective on TED (\cDC). The $p$ values for all other experiment pairs lie above $0.1$. From this we can conclude that it is indeed an acceptable practice to log deterministically.

\section{Analysis}
\label{sec:analysis}

\citet{LangfordETAL:08} show that counterfactual learning is impossible unless the logging system sufficiently explores the output space. This condition is seemingly not satisfied if the logging systems acts according to a deterministic policy. Furthermore, since techniques such as ``exploration over time'' \cite{StrehlETAL:10} are not applicable to commercial SMT systems that are not frequently changed over time, the case of counterfactual learning for SMT seems hopeless. However, our experiments present evidence to the contrary. In the following, we present an analysis that aims to explain this apparent contradiction.

\paragraph{Implicit Exploration.}

In an experimental comparison between stochastic and deterministic logging for bandit learning in computational advertising, \citet{ChapelleLi:11} observed that varying contexts (representing user and page visited) induces enough exploration into ad selection such that learning becomes possible. A similar implicit exploration can also be attributed to the case of SMT: An identical input word or phrase can lead, depending on the other words and phrases in the input sentence, to different output words and phrases. Moreover, an identical output word or phrase can appear in different output sentences. Across the entire log, this implicitly performs the exploration on phrase translations that seems to be missing at first glance.

\paragraph{Smoothing by Multiplicative Control Variates.}

The DPM estimator can show a degenerate behavior in that the objective can be minimized simply by setting the probability of every logged data point to $1.0$. This over-represents logged data that received low rewards, which is undesired. Furthermore, systems optimized with this objective cannot properly discriminate between the translations in the output space. This can be seen as a case of translation invariance of the objective, as has been previously noted by \citet{SwaminathanJoachimsNIPS:15}: Adding a small constant $c$ to the probability of every data point in the log increases the overall value of the objective without improving the discriminative power between high-reward and low-reward translations.

DPM+R solves the degeneracy of DPM by defining a probability distribution over the logged data by reweighting via the multiplicative control variate. After reweighting, the objective value will decrease if the probability of a low-reward translation increased, as it takes away probability mass from other, higher reward samples. Because of this trade-off, balancing the probabilities over low-reward and high-reward samples becomes important, as desired.

\paragraph{Smoothing by Additive Control Variates.}

Despite reweighting, DPM+R can still show a degenerate behavior by setting the probabilities of only the highest-reward samples to $1.0$, while avoiding all other logged data points. This clearly hampers the generalization ability of the model since inputs that have been avoided in training will not receive a proper ranking of their translations.

The use of an additive control variate can solve this problem by using a reward estimate that takes the full output space into account. The objective will now be increased if the probability of translations with high estimated reward is increased, even if they were not seen in training. This will shift probability mass to unseen data with high estimated-reward, and thus improve the generalization ability of the model.

\section{Conclusion}

In this paper, we showed that off-policy learning from deterministic bandit logs for SMT is possible if smoothing techniques based on control variates are used. These techniques will avoid degenerate behavior in learning and improve generalization of empirical risk minimization over logged data. Furthermore, we showed that standard off-policy evaluation is applicable to SMT under stochastic logging policies. 

To our knowledge, this is the first application of counterfactual learning to a complex structured prediction problem like SMT. Since our objectives are agnostic of the choice of the underlying model $\pi_w$, it is also possible to transfer our techniques to non-linear models such as neural machine translation. This will be a desideratum for future work.

\section*{Acknowledgments}
The research reported in this paper was supported in part by the German research foundation (DFG), and in part by a research cooperation grant with the Amazon Development Center Germany.

\bibliography{emnlp2017}
\bibliographystyle{emnlp_natbib}

\end{document}